%% file: main.tex
\PassOptionsToPackage{table}{xcolor} 
\documentclass[10pt,twocolumn,letterpaper]{article}

\usepackage{iccv}              


\definecolor{iccvblue}{rgb}{0.21,0.49,0.74}

\usepackage[pagebackref,breaklinks,colorlinks,allcolors=iccvblue]{hyperref}
\usepackage{physics, amsmath, bm}
\usepackage{microtype}
\usepackage{graphicx}
\usepackage{booktabs} 
\usepackage{multirow}
\usepackage[table]{xcolor}

\usepackage{amsmath}
\usepackage{dsfont}
\usepackage{float}
\usepackage{multirow}
\usepackage{multicol}
\usepackage{tablefootnote}
\usepackage{pdflscape}
\usepackage{pifont}
\usepackage{subcaption} 
\usepackage{booktabs}
\usepackage[accsupp]{axessibility}
\definecolor{columbiablue}{rgb}{0.8, 1, 1.0}
\definecolor{bisque}{rgb}{1.0, 0.89, 0.77}
\definecolor{customblue}{RGB}{173, 216, 230} 
\def\paperID{11338} 
\def\confName{ICCV}
\def\confYear{2025}

\title{SplatTalk: 3D VQA with Gaussian Splatting}

\author{
    Anh Thai$^{1,2}$\textsuperscript{*} \quad 
    Songyou Peng$^{2}$ \quad 
    Kyle Genova$^{2}$ \quad 
    Leonidas Guibas$^{2}$ \quad 
    Thomas Funkhouser$^{2}$ \\\\[-0.3cm]
    $^1$Georgia Institute of Technology \quad \quad 
    $^2$Google DeepMind \\
    {\small \textsuperscript{*}Work done as a student researcher at Google DeepMind.}
}

\begin{document}
\twocolumn[{%
\renewcommand\twocolumn[1][]{#1}%
\maketitle
\footnotetext[1]{Work done as a student researcher at Google DeepMind.}

\begin{center}
    \centering
    \captionsetup{type=figure}
    \vspace{-3ex}
    \includegraphics[width=\textwidth]{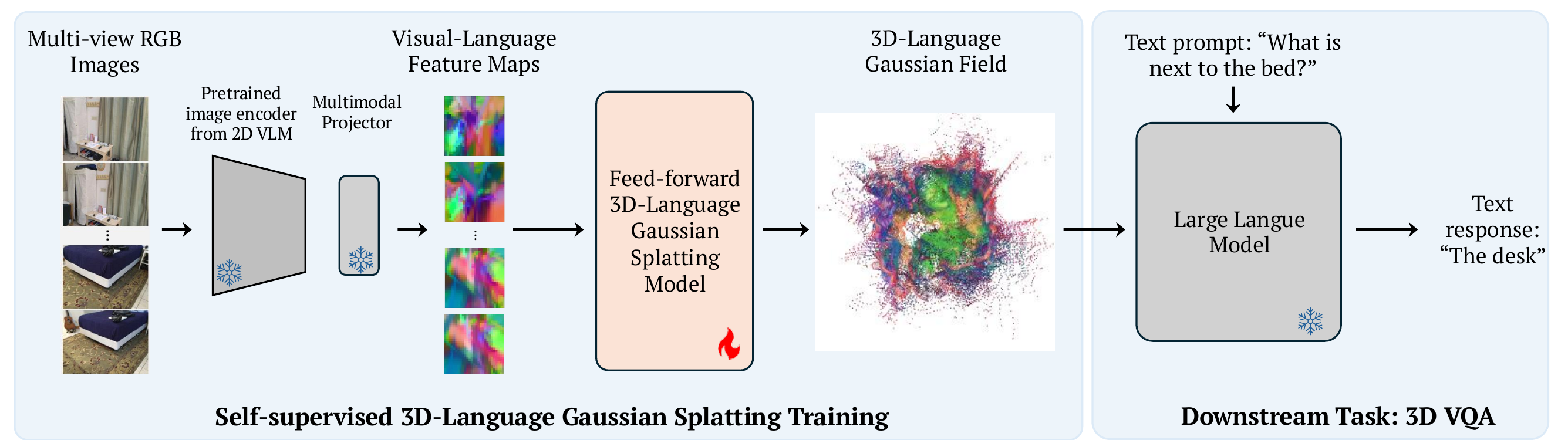}
    \vspace{-3.6ex}
    \captionof{figure}{\textbf{SplatTalk.} We propose a self-supervised 3D-Language Gaussian Splatting model trained from multi-view RGB images. First, images are encoded using a pretrained 2D Vision-Language Model (VLM) and projected into visual-language feature maps via a multimodal projector. These feature maps are then learned within a feed-forward 3D-Language Gaussian Splatting model, producing a 3D-Language Gaussian Field that encodes spatial and semantic information in 3D space. During inference, the 3D Gaussian features are directly queried by a Large Language Model (LLM) to perform 3D question-answering (3D VQA) tasks.}
\label{fig:teaser}
    
\end{center}%
}]
\input{sec/0_abstract}    
\input{sec/1_intro_new}
\input{sec/2_related_work}
\input{sec/3_method}

\input{sec/4_experiment}

\input{sec/5_conclusion}
\clearpage
\input{sec/6_supp_mat}
\clearpage

{
    \small
    \bibliographystyle{ieeenat_fullname}
    \bibliography{main}
}



\end{document}

%% file: sec/0_abstract.tex
\begin{abstract}
Language-guided 3D scene understanding is important for advancing applications in robotics, AR/VR, and human-computer interaction, enabling models to comprehend and interact with 3D environments through natural language. While 2D vision-language models (VLMs) have achieved remarkable success in 2D VQA tasks, progress in the 3D domain has been significantly slower due to the complexity of 3D data and the high cost of manual annotations. In this work, we introduce SplatTalk, a novel method that uses a generalizable 3D Gaussian Splatting (3DGS) framework to produce 3D tokens suitable for direct input into a pretrained LLM, enabling effective zero-shot 3D visual question answering (3D VQA) for scenes with only posed images. During experiments on multiple benchmarks, our approach outperforms both 3D models trained specifically for the task and previous 2D-LMM-based models utilizing only images (our setting), while achieving competitive performance with state-of-the-art 3D LMMs that additionally utilize 3D inputs. Project website: \url{https://splat-talk.github.io/}
\end{abstract}

%% file: sec/1_intro_new.tex
\section{Introduction}
\label{sec:intro}

The rise of Large Multimodal Models (LMMs), following the success of Large Language Models (LLMs), has become increasingly significant in a world where visual understanding is crucial. While 2D LMMs have seen rapid improvements, enabling models to analyze images and answer complex questions~\cite{wang2025cogvlm,liu2023visual,li2024llava,rasheed2024glamm}, progress in 3D LMMs has been slower. This gap is largely due to the scarcity of 3D data compared to 2D datasets. Further, the availability of language-annotated 3D data is even more limited, as annotating 3D scenes since it is significantly more complex and expensive than labeling 2D images. These challenges have hindered the development of language-guided 3D reasoning, making 3D LMMs an important yet underexplored area in multimodal research. In this work, we focus on 3D Visual Question Answering (3D VQA) as it provides a strong benchmark for evaluating a model’s ability to understand and reason about 3D scenes.

Recently, significant progress has been made in 3D multimodal learning, particularly for indoor scenes~\cite{huang2023embodied,hong20233d,linghu2025multi,zhu2024llava}. Several approaches rely on posed images and 3D inputs, like meshes \cite{wang20233d}, point clouds~\cite{chen2024grounded,huang2023embodied,peng2023openscene} and voxels~\cite{fu2024scene}, to associate 3D geometry with language features.  As such, they have limited applicability in scenarios where a full 3D surface reconstruction is not available.

Previous 
works have explored integrating semantic features into NeRF \cite{kerr2023lerf} or 3D Gaussian Splatting (3DGS)~\cite{qin2024langsplat}, which rely only upon RGB image inputs.   However, research on effectively associating 3D geometry with language fetures remains limited.   Some methods align individual 3D locations with language features via implicit functions optimized individually for every scene \cite{kerr2023lerf,engelmann2024opennerf}.   Others employ SAM~\cite{kirillov2023segment} to first segment objects and object parts, followed by CLIP~\cite{radford2021learning} to extract object-centric features from the generated masks. These approaches allow models to query information about individual points and/or objects. However, it disregards spatial relationships between objects, making it unsuitable for 3D VQA spatial reasoning tasks.  


More recent work ~\cite{li2024llava} has demonstrated the power of LLMs to answer VQA questions using only a sampling of tokens extracted from images, without any underlying 3D representation.   While this approach is inspiring, it relies upon the LLM to make sense of a set of tokens that are arranged according to images rather than elements in the 3D scene.   As a result, it struggles to answer queries about specific 3D locations (e.g., ``What is here?") and questions requiring a holistic understanding of the entire 3D scene, where the answer depends on establishing correspondences between objects across multiple images (e.g., ``What is opposite the door at the other end of the room"?).   In this work, we aim to address these issues by consolidating tokens in 3D before providing them to the LLM.  

Our approach is called \emph{SplatTalk}.  Given a set of posed RGB images, we integrate language features extracted from the RGB images to build a 3D Gaussian Splatting representation, from which 3D tokens can be extracted for direct input into a pretrained LLM to answer challenging 3D VQA queries (see Fig.~\ref{fig:teaser}).  The main research contributions are the new methods for 1) integrating language features into the Gaussian representation, and 2) extracting tokens that can be provided directly to an LLM.   These methods avoid the issues of concurrent work, ChatSplat ~\cite{chen2024chatsplat},  which produces tokens that are not directly compatible with an LLM, making language-based reasoning more difficult.

Overall, SplatTalk has a unique combination of three advantages.  First, it does not require any 3D surface representation as input.  Second, it leverages the full power of a pre-trained LLM to answer difficult 3D VQA questions.  Third, it produces a concise 3D representation that can be queried and/or edited in 3D (e.g., cropped to a 3D region, rendered from a novel view, etc.).   During experiments with 3D VQA benchmarks, we find that it outperforms 3D models trained only for the benchmark tasks, demonstrating the value of utilizing pre-trained LLMs.  SplatTalk further outperforms prior image-only methods using LLMs, showing the value of building a 3D scene representation.   Finally, it even achieves competitive performance with SOTA 3D LMMs that utilize 3D scene representation inputs.



In summary, our key contributions are:
\begin{itemize}
    \item We introduce the first self-supervised 3D Gaussian-based method for large-scale zero-shot 3D VQA, SplatTalk, requiring only multi-view images without explicit depth, point clouds or 3D-language annotated supervision.
    \item We show that mean feature extraction from 3D Gaussians encodes holistic scene concepts for 3D VQA, supported by theoretical justification.
    \item We introduce entropy-adaptive token sampling, leading to better 3D VQA performance without additional training.
    \item We demonstrate that SplatTalk outperforms 2D LMMs and achieves performance on par with 3D models using point clouds on various 3D VQA benchmarks, demonstrating the effectiveness of our 3D-aware representations.
\end{itemize}

%% file: sec/2_related_work.tex
\section{Related Work}
\label{sec:related_work}
\input{figure_tex/arch}

\textbf{2D LMMs.}  With advancements in large language models (LLMs), the ability of models to comprehend images and respond to language prompts related to their content has improved significantly. Notably, SOTA models~\cite{openai2023gpt4v,geminiteam2024geminifamilyhighlycapable, anthropic2024claude3,lu2024deepseekvl,li2024llava} have demonstrated remarkable capabilities in processing multiple images of a 3D scene and answering questions about the shared content across these images. Although multiple RGB images inherently contain 3D information, they do not explicitly model 3D understanding, potentially leading to sub-optimal performance on tasks that require explicit 3D reasoning. In this work, we build upon the strong performance of 2D vision-language models (2D-VLMs), particularly LLaVA-OneVision~\cite{li2024llava}, and integrate them with the 3D priors provided by the 3DGS framework to improve 3D spatial understanding capabilities.

\noindent\textbf{3D LMMs.} Recent advancements in 3D indoor scene understanding have explored integrating 3D priors into LLMs using inputs such as 3D point clouds~\cite{huang2023embodied,chen2024ll3da} or multi-view RGB images combined with depth or 3D point clouds~\cite{zhu2024llava}. These approaches employ various strategies to embed 3D structural information and language features into a shared representation space. While most methods~\cite{zhu2024llava,huang2023embodied} fine-tune pretrained visual encoders (both 2D and 3D) alongside LoRA parameters or even the full set of LLM parameters, others~\cite{li2023blip,kang2025robin3dimproving3dlarge} opt for a more lightweight approach, training only a projection layer to map visual features into the language feature space. In contrast, our method does not require any explicit depth or 3D inputs. Instead, we leverage the 3DGS framework, which has shown strong capabilities in representing 3D scenes using only multi-view RGB images.

\noindent\textbf{Generalizable 3D Gaussian Splatting.} This body of work explores generalizable 3DGS as an alternative to the per-scene optimization paradigm used in traditional 3DGS methods, aiming to improve novel view synthesis across unseen scenes. Our method builds on FreeSplat~\cite{wang2025freesplat}, which is designed for sparse-view synthesis in indoor scenes. The method begins by extracting multi-scale features from input RGB images using CNN backbones. It then constructs adaptive cost volumes from neighboring views to predict depth maps. These predicted depth maps are used to back-project 2D CNN-encoded features into 3D Gaussian triplets ($\boldsymbol{\mu},\boldsymbol{\omega},\boldsymbol{f})$ where $\boldsymbol{\mu}$ denotes the Gaussian's center, $\boldsymbol{\omega}$ the weight, and $\boldsymbol{f}$ the latent feature. To integrate multi-view information, FreeSplat introduces a Pixel-wise Triplet Fusion (PTF) module that uses a lightweight GRU to align and fuse overlapping 3D Gaussians across views efficiently. Finally, the fused triplets are passed through an MLP to decode rendering parameters, $\boldsymbol{\Sigma}, \boldsymbol{\alpha},\boldsymbol{s}$.
Our approach extends the FreeSplat framework to enable the learning of generalizable 3D Gaussian language fields. 

Similar to FreeSplat~\cite{wang2025freesplat},~\cite{chen2024mvsplat,charatan2024pixelsplat} leverage multi-view stereo (MVS) cues to refine depth estimation and improve geometric consistency in 3D Gaussian representations. These methods primarily focus on RGB novel view synthesis from sparse inputs ($2–10$ views). Our work builds on the SOTA model, FreeSplat~\cite{wang2025freesplat}, but diverges significantly in both objectives and methodology. First, while previous methods focus on photorealistic RGB reconstruction, we aim at semantic 3D scene understanding. Second, instead of reconstructing only a localized portion of the scene as shown in these works, our goal is to represent entire 3D environments, capturing both spatial and semantic relationships holistically.

\noindent\textbf{Language-guided 3D Gaussian Splatting}. LangSplat~\cite{qin2024langsplat} is one of the first approaches that integrates semantic features into 3DGS. However, this method requires training on individual scenes, which is significantly expensive, and lacks generalization to unseen scenes. Similarly, follow-up works~\cite{wu2025opengaussian,li2024instancegaussian} explore 3DGS for open-vocabulary object segmentation but do not focus on reasoning tasks like 3D VQA. Further, they primarily rely on scene-specific training, limiting their scalability. In contrast, we propose a fully self-supervised framework that learns 3D-aware visual representations without requiring per-scene training. While some works~\cite{wang2024gsemsplatgeneralizablesemantic3d,fan2025large} explores a generalizable 3DGS and semantic modeling approach, they primarily focus on open-vocabulary segmentation, without addressing the broader challenges of free-form language reasoning required for 3D VQA.

ChatSplat~\cite{chen2024chatsplat} is a concurrent work that explores natural language conversation within the 3DGS framework. However, it optimizes on a per-scene basis and requires rendering 3D features into 2D feature maps during inference. Additionally, its training pipeline is not fully end-to-end, as it first trains the RGB Gaussians separately before incorporating language features. In contrast, our approach leverages scene latent features that encapsulate both RGB and semantic information, enabling the joint optimization of both modalities in a fully end-to-end manner.

%% file: figure_tex/arch.tex
\begin{figure*}[t!]
\centering
\includegraphics[width=\linewidth]{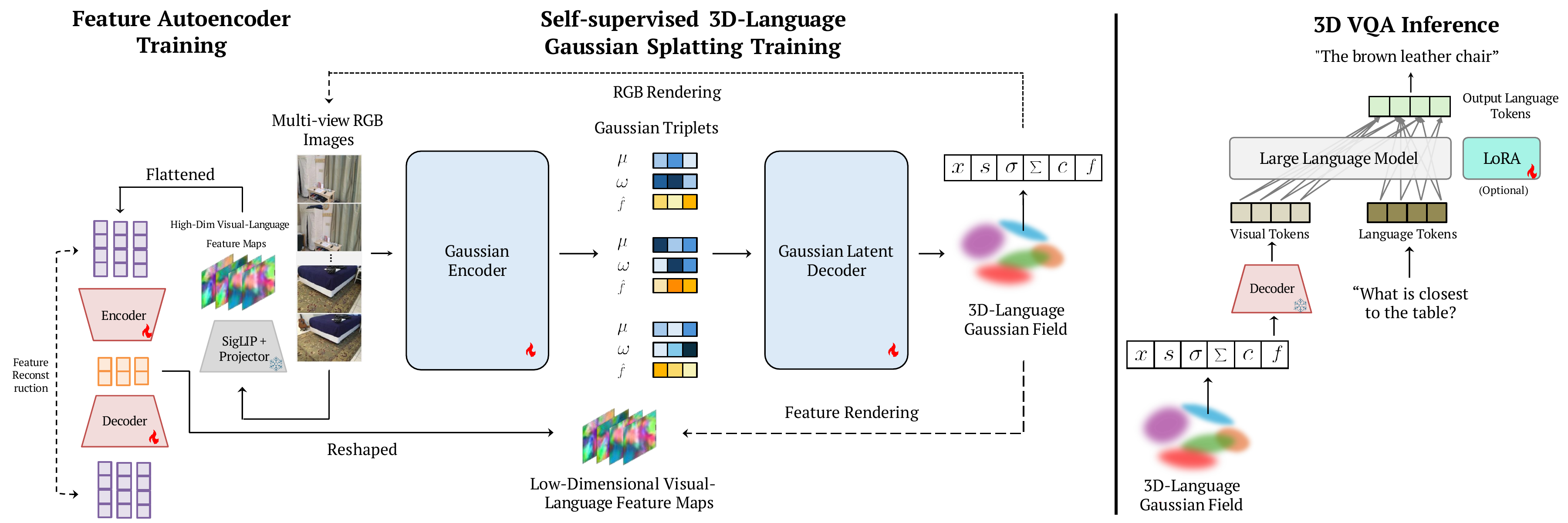}
\caption{\textbf{Overview.} \emph{Left}: During the self-supervised 3D-language Gaussian Splatting training phase, multiple RGB input views are first encoded into Gaussian latent features (Gaussian triplets). These latent features are then decoded into Gaussian parameters for rendering, along with a low-dimensional visual-language feature $f$.  To ensure proper supervision of this low-dimensional feature, we train an autoencoder that maps the high-dimensional, unbounded features obtained from LLaVA-OV, specifically, the visual tokens serving as direct inputs to the LLM, onto a low-dimensional hypersphere space. \emph{Right}: During 3D VQA inference, visual-language features are directly extracted from the 3D Gaussians. These features are then mapped back to the original high-dimensional space using the pretrained decoder and subsequently used as direct visual token inputs to the LLM. LoRA fine-tuning of the LLM is optional.}
\label{fig:arch}
\vspace{-15pt}
\end{figure*}

%% file: sec/3_method.tex
Our approach extends the FreeSplat framework to enable the learning of generalizable 3D Gaussian language fields. 


\section{Method}
\label{sec:method}
\input{figure_tex/example_viz}

The problem we address is formulated as follows: Given a sequence of \( N \) posed RGB images captured from a 3D indoor scene, along with language prompts related to the scene's content, the goal is to generate the corresponding language responses. That is, we address 3D VQA tasks with solely posed RGB images as inputs.

We provide an overview of our approach in Fig.~\ref{fig:arch}, which consists of three main steps: 1) Training the feature autoencoder, 2) Self-supervised 3D-Language Gaussian Splatting Training and 3) 3D VQA Inference. First, we extract the high-dimensional language features from LLaVA-OV~\cite{li2024llava} for each RGB image by using the pretrained image encoder and multimodal projector. Since these features are high-dimensional, unbounded and sparse, we train an autoencoder to project them into a more compact, low-dimensional space. These low-dimensional features then serve as pseudo ground-truth for training the language representations in SplatTalk. SplatTalk architecture is a feed-forward network consisting of a Gaussian encoder and decoder that predicts Gaussian parameters for rendering. During 3D VQA inference stage, we extract the trained language features from the 3D Gaussians and directly feed them into the LLM, along with the language prompt, to generate free-form responses.


\subsection{Integrating Language Features}

An important step of our computational pipeline is the self-supervised method to integrate language features into a generalizable 3DGS framework.   
We identify four key factors that shape our approach for this step: (1) Leveraging features that align with the LLM’s input space for direct language reasoning, (2) Efficient feature dimensionality reduction for effective CUDA-based rendering, and (3) Joint training of RGB and semantic features for coherent multimodal representation learning.



\noindent\textbf{Visual Tokens for 2D Pseudo Ground-truth Features.} We extract the visual tokens used as inputs to the LLM for training our 3DGS framework. Our empirical findings indicate that extracting features directly from the visual encoder for training 3DGS requires retraining the multimodal projector layer which aligns the visual with language spaces. This observation is further corroborated by~\cite{chen2024chatsplat}, which demonstrates that features extracted before the multimodal projector cannot be seamlessly integrated into the language model. A possible explanation is that the image encoder's native feature space (before the projector) is high-dimensional and unbounded, leading to sparser features with greater variance across different dimensions. As a result, the projector becomes more sensitive to subtle variations in the input features, making it difficult to directly map them into a token representation that aligns with the LLM's understanding. In contrast, the visual tokens generated after the projector are already well-aligned with the language model’s latent space, allowing the LLM to interpret and reason with them directly.
\input{table/scanqa_sqa}

\noindent\textbf{Feature Dimensionality Reduction.} Building on the previous factor, directly optimizing these language embeddings within a differentiable rendering pipeline is challenging because they are extremely high-dimensional (e.g., $3584$ dimensions), sparse, and contain unbounded values, making learning both unstable and impractical. Therefore, we learn a generalizable autoencoder that projects these sparse, high-dimensional visual tokens into a compact unit hypersphere with much lower dimension (e.g., $256$ dimensions), ensuring feature smoothness. Note that the language representations from LLaVA, which support reasoning and free-form language comprehension, are significantly more complex than those from CLIP or segmentation-based models like SAM. Different from prior works that compress semantic features into extremely low-dimensional vectors (e.g., $3–16$ dimensions~\cite{qin2024langsplat,wu2025opengaussian}), which simplifies rendering but results in significant information loss, our approach prioritizes preserving rich language features by maintaining a higher-dimensional representation (e.g., $256$). To efficiently render high-dimensional feature maps on CUDA, we draw inspiration from~\cite{zhou2024feature} and implement a parallel CUDA rasterizer pipeline, where semantic features and RGB are rendered simultaneously, using shared Gaussian parameters.

\input{table/msqa}

Unlike previous methods~\cite{qin2024langsplat,chen2024chatsplat} that train a separate autoencoder for each individual scene, we train a single autoencoder across all training scenes, each consisting of multiple frames. This approach improves the autoencoder's generalizability, reducing computational overhead by eliminating the need for retraining to compress features from unseen scenes. Training the autoencoder separately from the 3DGS model ensures that it is dedicated solely to compressing high-dimensional feature vectors while preserving the essential scene information without being influenced by the complexities of the rendering pipeline. Our empirical findings further indicate that jointly training a feature decoder to project low-dimensional embeddings back to the high-dimensional space introduces significant instabilities in the training process.


\noindent\textbf{Joint-training RGB and Language.} In contrast to previous approaches that first train RGB Gaussians and then freeze them before integrating language features~\cite{chen2024chatsplat,qin2024langsplat}, we adopt a joint training strategy, where RGB and language features are learned simultaneously. This is due to the fact that visual RGB and semantic features play a complementary role in effectively representing 3D scenes~\cite{peng20243d}. In feedforward models like FreeSplat, where the scene is first encoded into a Gaussian latent space before being decoded into specific Gaussian parameters for RGB rendering, these Gaussian latent features inherently capture holistic scene information, including both appearance (RGB) and semantics. Based on this observation, we introduce a new Gaussian parameter decoder head that predicts semantic features for each 3D Gaussian alongside rendering parameters, enabling joint optimization. Unlike approaches that explicitly model these interactions~\cite{peng20243d,wang2024lift3d} using extra cross-attention or projection layers, SplatTalk implicitly integrates semantic and visual information within the Gaussian representation. This paradigm 
naturally facilitates multimodal interactions between RGB and language features. 

To incorporate more views and comprehensively cover the scene, we reduce the input resolution of the RGB images. This adjustment aligns well with the inherently low spatial resolution of the visual tokens extracted from LLaVA-OV, and also prevents the model from disproportionately allocating resources to modeling RGB features, thus encouraging a balanced representation of semantic and RGB information. 

These design choices establish our approach as a fundamentally novel method for integrating language features into 3DGS, grounded in first-principles observations rather than heuristic adaptations. By addressing the core challenges of multimodal alignment, feature compression, and joint optimization, our approach is not merely a trivial extension of existing works, but a carefully designed framework that ensures both effectiveness and stability in learning rich, spatially-aware semantic representations.

\subsection{Extracting 3D Language Features}\label{sec:extract}
During inference, we directly extract language features at the mean position of each 3D Gaussian. Although the rendering process incorporates each Gaussian’s covariance and opacity when projecting 3D features to 2D, we demonstrate that \textit{scene semantics can be effectively captured by the mean feature of each Gaussian, $\bm{f}_i$, alone, without relying on these additional factors.} This is supported by the conceptual similarity between our training optimization process and the Expectation-Maximization (EM) algorithm. For simplification, let $\mathcal{L}=\sum_{t=1}^T\Vert\bm{F}_t(x)-\bm{F}^{gt}_t(x)\Vert^2$ be the loss, where $\bm{F}_t(x)$ is the predicted rendered feature map of frame $t$ at pixel $x$ and $\bm{F}^{gt}$ is the pseudo ground-truth feature map obtained from LLaVA-OV. Replacing $\bm{F}_t(x)$ by the rendering equation and solving for the optimal $\bm{f}^*_i$, we obtain
$
\bm{f}_i^*=\frac{\sum_{t,x}R_i(t,x)\bm{F}^{gt}_t(x)}{\sum_{t,x}R_i(t,x)}
$
where $R_{i}(t,x)=\frac{\alpha_i\mathcal{N}(x;\bm{\mu}_i,\bm{\sum}_i)}{\sum_j\alpha_j\mathcal{N}(x;\bm{\mu}_j,\bm{\sum}_j)}$ is the Gaussian's contribution to pixel $x$ at viewpoint $t$. This mirrors the EM procedure: estimating $R$ by rendering (E-step) and updating $f_i$ by minimizing reconstruction loss (M-step).
Consequently, the optimal 3D feature $\bm{f}_i^*$ for each Gaussian is a weighted sum of the 2D feature maps,
and the set of 3D Gaussian features $\{\bm{f}_i\}_{i=1}^N$ encodes the scene's semantic holistically from $T$ 2D inputs. These features are then projected back into the original language embedding space using the pretrained decoder and directly provided as inputs to the LLM for downstream 3D-language tasks.

In open-vocabulary segmentation, querying Gaussian centers is effective~\cite{bhalgat2024n2f2, qin2024langsplat, wu2025opengaussian} because each Gaussian is assumed to represent a distinct, meaningful object or region. These methods use object-centric embeddings (e.g., SAM-masked CLIP or LSeg) and retrieve objects via cosine similarity to text queries. In contrast, 3D reasoning tasks lack this one-to-one correspondence. Each Gaussian encodes parts of a scene-level representation, where meaning stems from spatial and relational structure. Querying single Gaussians is thus insufficient for context-dependent reasoning. While our method also queries at Gaussian means, this is not trivial; our EM-based formulation offers a principled justification.

\subsection{3D VQA Inference}\label{sec:entropy_sampling}
\noindent\textbf{Entropy Adaptive Gaussian Sampling.}
SplatTalk leverages triplet fusion adopted in FreeSplat~\cite{wang2025freesplat}, which reduces the number of Gaussians. However, complex scenes still require significantly more Gaussians than the LLM’s visual token limit, and not all are equally informative. To address this, we propose entropy-based sampling: selecting the top $k$ Gaussians with the highest language feature entropy, where $k$ matches the LLM’s visual token capacity.

\noindent\textbf{3D Tokens as Visual Inputs.} After 3DGS training, each 3D Gaussian encodes a feature that is a weighted aggregation of 2D semantic information across all input views (Sec.~\ref{sec:extract}). These features are also spatially grounded in their 3D coordinates. Because each token corresponds to a specific location in 3D space, its position is already implicitly encoded in the feature through the rendering and fusion process. Unlike 2D patches in a fixed grid, these 3D tokens form an unordered set, similar to a point cloud where spatial relationships are captured by the features themselves, not by sequence position. Therefore, we do not require explicit positional encodings or a fixed input order. The 3D tokens, extracted by sampling features at each Gaussian’s mean, are directly fed into the LLM.



\subsection{Optional - Finetuning on 3D VQA Datasets}
While SplatTalk is capable of performing the 3D VQA task in a zero-shot manner, it also allows for flexible fine-tuning on 3D VQA datasets to enhance alignment and reasoning capabilities. Rather than training the model end-to-end, which is both computationally expensive and prone to overfitting on limited datasets, we optimize only the LoRA parameters applied to the LLM during fine-tuning.

%% file: figure_tex/example_viz.tex
\begin{figure*}[t!]
\centering
\includegraphics[width=\linewidth]{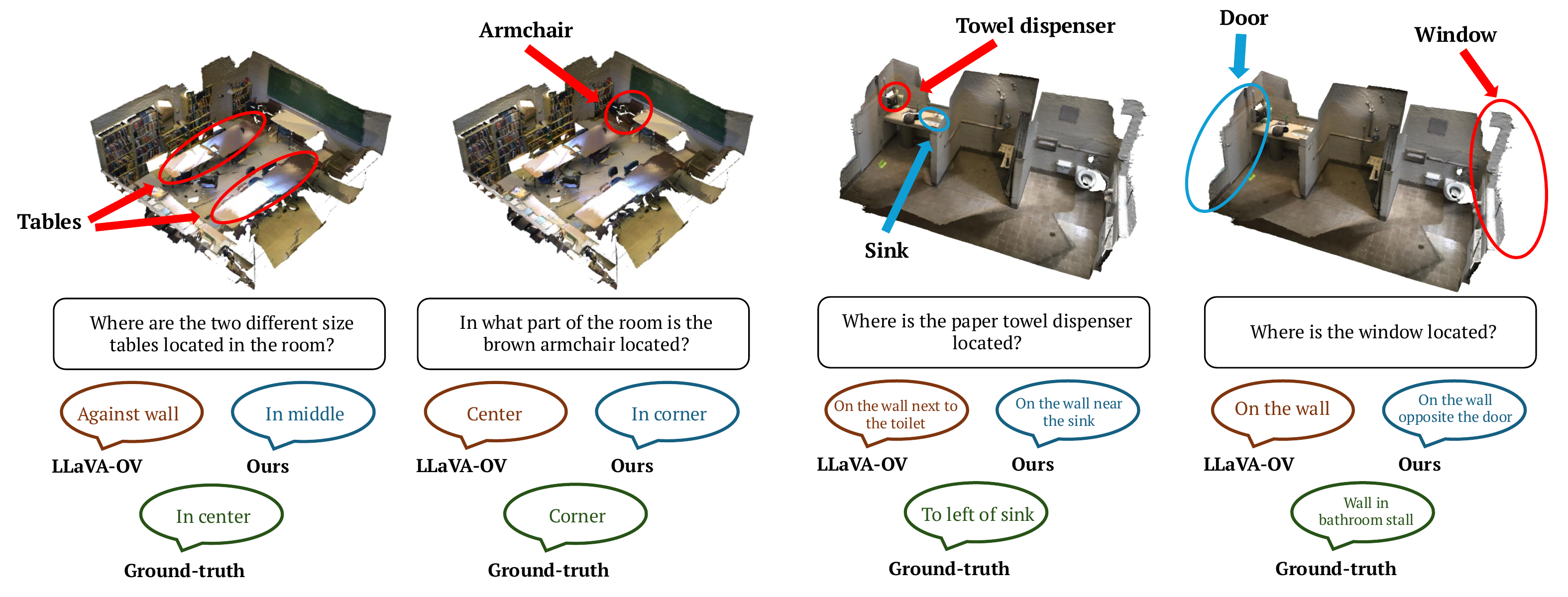}
\caption{\textbf{Qualitative results on ScanQA}, scene0030\_00 and scene0084\_00. We compare responses from LLaVA-OV, our model (Ours), and the ground truth (GT) for spatial reasoning questions in 3D VQA. Each scene highlights the referenced objects with red circles, and key relative objects are marked in blue. The answers from each model are displayed in color-coded speech bubbles: LLaVA-OV (brown), Ours (blue), and GT (green). With its 3D-aware representation, our model exhibits improved spatial reasoning capabilities, accurately identifying relationships between objects that are far apart and may never co-occur in the same image (e.g., door and window). More examples in Supp.}
\label{fig:example_viz}
\vspace{-15pt}
\end{figure*}

%% file: table/scanqa_sqa.tex
\begin{table*}[t!]
\begin{center}

\begingroup
\setlength{\tabcolsep}{4pt} 
\renewcommand{\arraystretch}{1.2} 
\caption{\textbf{Evaluation of 3D VQA on ScanQA~\cite{azuma2022scanqa} and SQA3D~\cite{ma2022sqa3d} datasets.} We compare SplatTalk against various baselines, including 3D LMMs, which take point clouds (PC) or both point clouds and images (PC+I) as visual inputs, and 2D LMM-based models, which rely solely on images (I). As a 2D LMM-based approach, SplatTalk outperforms other 2D LMM-based models, surpasses some 3D LMMs, and achieves competitive performance against the strongest 3D LMM baselines. Our models are highlighted in blue, bold indicates best performance among the 2D-LMM methods.}
\scalebox{0.83}{

\begin{tabular}{lcccccccc}
\hline
\toprule
 \multirow{2}{*}{Method} & \multirow{2}{*}{Modality} & \multicolumn{5}{c}{ScanQA (val)}  & \multicolumn{2}{c}{SQA3D (test)}\\
\cmidrule(lr){3-7} \cmidrule(lr){8-9}
& & CIDEr &  METEOR & ROUGE & EM@1 & EM@1-R &  EM@1 & EM@1-R\\
\midrule
\multicolumn{9}{l}{\textbf{\textit{Specialist Models}}} \\
ScanQA~\cite{azuma2022scanqa} & PC & 64.9 &  13.1 & 33.1 & 21.1 & -& 47.2 & -\\
ScanRefer+MCan~\cite{yu2019deep}  & PC & 55.4 & 11.5 & 30.0 & 18.6 &- & -  & -\\
3D-VisTA~\cite{zhu20233d} & PC&69.6 & 13.9 & 35.7 & 22.4 & - & 48.5 & -\\
SQA3D~\cite{ma2022sqa3d} & PC & - &- &- &- &- & 46.6 & - \\
\midrule

\multicolumn{9}{l}{\textbf{\textit{Generalist 3D LMMs}}} \\
Chat-Scene~\cite{huang2023chat} & PC+I &87.7 & 18.0 & 41.6 & 21.6 &- & 54.6 & 57.5 \\
Scene-LLM~\cite{fu2024scene} & PC+I & 80.0 & 15.8 & 35.9 & - & - & 53.6 & -\\
Grounded 3D-LLM~\cite{chen2024grounded} & PC & 72.7 & - & - & - & - & - & -\\
Chat-3D-v2~\cite{wang2023chat} & PC & 77.1 & 16.1 & 40.1  & - & - & 54.7 & -\\
\midrule
\multicolumn{9}{l}{\textbf{\textit{Finetuned 3D LMMs}}}\\
3D-LLM~\cite{hong20233d} & PC+I & 69.4 &  14.5 & 35.7 & 20.5 & - & - & - \\

LEO~\cite{huang2023embodied} & PC+I &101.4 &  20.0 & 49.2 & 24.5 & 47.6 & 50.0 & 52.4\\
Scene-LLM~\cite{fu2024scene} & PC+I & 80.0 &  16.6 & 40.0 & 27.2 & - & 54.2 & -\\
LL3DA~\cite{chen2024ll3da} & PC & 76.8 & 15.9 & 37.3 & - & - & - & - \\
Chat-3D~\cite{wang2023chat} & PC & 53.2 & 11.9 & 28.5 & - & - & - & - \\
\midrule

 \multicolumn{9}{l}{\textbf{\textit{2D LMM-Based}}} \\
 VideoChat2~\cite{li2024mvbench} & I & 49.2 & 9.5 & 28.2 & 19.2 & - & 37.3 & - \\
 GPT-4V~\cite{openai2023gpt4v} & I & 59.6 & 13.5 & 33.4 & - & - & - & - \\
 Claude~\cite{anthropic2024claude3} & I & 57.7 & 10.0 & 29.3 & - & - & - & -  \\
 LLaVA-NeXT-Video~\cite{li2024llavanextinterleavetacklingmultiimagevideo} & I & 46.2 & 9.1 & 27.8 & - & - & 34.2 & - \\
 LLaVA-OV~\cite{li2024llava} & I & 50.0 & 13.0 & 29.4 & 15.6 & 32.5 & 25.4 & 35.8\\
 \rowcolor{columbiablue} SplatTalk & I & 61.7 & 14.2 & 32.7 & 17.1 & 32.2 & 26.1 & 41.3 \\
 \rowcolor{columbiablue} SplatTalk-ScanQA-FT & I & 77.1 & 15.4 & \textbf{38.7} & 22.3 & \textbf{38.3} & 42.0 & 44.7\\
 \rowcolor{columbiablue} SplatTalk-3DVQA-FT & I & \textbf{77.5} & \textbf{15.6} & 38.5 & \textbf{22.4} & \textbf{38.3} &  \textbf{47.6} & \textbf{49.4} \\
 \bottomrule
\end{tabular}
\label{tab:scanqa_sqa3d}
}
\endgroup

\vspace{-20pt}
\end{center}

\end{table*}

%% file: table/msqa.tex
\begin{table*}[t!]
\begin{center}

\begingroup
\setlength{\tabcolsep}{4pt} 
\renewcommand{\arraystretch}{1.2} 
\caption{\textbf{Evaluation of 3D VQA on MSR3D~\cite{linghu2025multi}}. We compare SplatTalk against various baselines, all evaluated in a zero-shot setting. T denotes models that rely solely on text inputs without interleaving images, PC indicates models that use point cloud as the visual input, and I represents models that process only image-based inputs. Our approach, highlighted in blue, outperforms all baselines.}
\scalebox{0.83}{

 \begin{tabular}{l|c c c|c c c c c c c}
        \toprule
        Model  & Input & Scene & Setting & Counting & Existence & Attributes & Spatial & Navigation & Others & Overall \\
        \midrule
        LEO~\cite{huang2023embodied}  & T & PC & zero-shot  & 0.80  & 15.5  & 11.8  & 7.3  & 2.3  & 15.3  & 7.8  \\
        LLaVA-OV~\cite{li2024llava} & T & I & zero-shot & 18.6 & 31.2 & 24.8 & 19.5 & 16.7 & 36.3 & 24.0\\
        \rowcolor{columbiablue} SplatTalk & T & I & zero-shot  & 19.6  & 60.3  & \textbf{44.0}  & \textbf{35.8}  & \textbf{35.5} & \textbf{61.8}  & \textbf{41.8}  \\
        \rowcolor{columbiablue} SplatTalk-ScanQA-FT& T & I & zero-shot  & \textbf{28.9}  & \textbf{66.0}  & 43.2  & 28.3  & 33.8  & 60.0  & 41.5  \\
        \rowcolor{columbiablue} SplatTalk-3DVQA-FT &  T & I & zero-shot & 24.3 & 57.0 & 36.1 & 7.8 & 16.4 & 35.0 & 26.5 \\
\bottomrule 
\end{tabular}
\label{tab:msqa}

}
\endgroup

\vspace{-20pt}
\end{center}

\end{table*}


%% file: sec/4_experiment.tex
\section{Experiments}
\label{sec:experiments}

\subsection{Implementation Details} 
We train SplatTalk with 500 scenes from the ScanQA training set. 
The initial self-supervised stage focuses purely on learning the 3D scene representation, without leveraging any text-annotated information.
For each scene, we uniformly sample 100 views around the scene and extract their per-patch features by passing them through the visual encoder of LLaVA-OV followed by the multi-modal projector. 
The subsequent SplatTalk model is trained on a single NVIDIA H100 (80GB) GPU. Both the LoRA fine-tuning and inference are also conducted on a single H100. More details can be found in Supp. Material.

\subsection{Datasets and Metrics}
\textbf{Evaluation Dataset}. We evaluate our method on ScanNet-based QA datasets: ScanQA~\cite{azuma2022scanqa}, SQA3D~\cite{ma2022sqa3d}, and MSR3D~\cite{linghu2025multi}. ScanQA is a benchmark designed for question answering in 3D indoor scenes, containing detailed spatial and semantic information extracted from ScanNet. SQA3D and MSR3D further extend this by introducing scenarios that describe an embodied agent's 3D position and state.

\noindent\textbf{Metrics}. For ScanQA and SQA3D, we evaluate performance using standard n-gram-based metrics, specifically CIDEr~\cite{vedantam2015cider}, METEOR~\cite{banerjee2005meteor}, ROUGE~\cite{lin2004rouge}, EM@1~\cite{goyal2017making}, and EM@1-Refined for ScanQA, and EM@1 and EM@1-Refined for SQA3D. In contrast, for MSR3D, we utilize the LLM-Match score~\cite{Liu2024MMBench} as recommended by the authors.

\subsection{Performance on 3D Question Answering Tasks}
\textbf{ScanQA}. We evaluate SplatTalk’s performance against several model families, each representing different levels of specialization and generalization. ``Specialist Models" refer to models specifically trained to solve ScanQA without exposure to other tasks. ``Generalist 3D LMMs" consist of models trained across a diverse range of 3D vision-language tasks, enabling broader generalization. ``Finetuned 3D LMMs" are models that have been further fine-tuned on the ScanQA training set to optimize performance for this particular dataset. Lastly, ``2D LMM-Based Models" include models that process only image inputs without leveraging depth or explicit 3D information. SplatTalk and its fine-tuned variants fall within this category.

We evaluate three different variants of our model on this task. The base model (SplatTalk) is trained solely to encode the semantic representation of the scene from multiple RGB images using 3DGS, without exposure to question-answering tasks. Note that this model is trained entirely in a self-supervised manner, without any access to language supervision or textual annotations during training. To further adapt it to downstream 3D VQA tasks, we fine-tune this base model on the ScanQA training set, referring to it as SplatTalk-ScanQA-FT, and on the combined training sets of ScanQA and SQA3D, denoting as SplatTalk-3DVQA-FT. We present results in Table~\ref{tab:scanqa_sqa3d}.

Compared to LLaVA-OV, whose 2D language feature maps are used to train SplatTalk, our base model, which benefits from a more 3D-aware representation due to the 3D Gaussian training process, achieves significantly higher performance (e.g. $61.7$ vs $50.0$ on CIDEr). Our language fine-tuned model variants, SplatTalk-ScanQA-FT and SplatTalk-3DVQA-FT, improve performance further. Notably, despite not being trained with depth supervision or explicit 3D information, our models achieve comparable results to both Generalist 3D LMMs and Finetuned 3D LMMs, while outperforming Specialist Models. Additionally, among 2D LMM-Based approaches, our base model without being fine-tuned with 3D VQA datasets outperforms other methods, with significantly higher performance on some metrics, including both open-source and closed-source models. We show qualitative results in Fig.~\ref{fig:example_viz}.

\input{table/combine_ablation}

\noindent\textbf{SQA3D}. Similar to our evaluation on ScanQA, we compare our model against both 3D LMMs and 2D LMMs. Compared to 2D LLaVA-OV features, our base model with 3D-aware features achieves a significant performance gain, highlighting the advantage of incorporating 3D information via 3DGS which aligns with our findings on ScanQA. Further fine-tuning on ScanQA alone and on the combined ScanQA + SQA3D dataset leads to substantial improvements. Notably, SplatTalk-3DVQA-FT surpasses specialist models that leverage 3D point cloud inputs, demonstrating the strength of our approach in 3D scene understanding.



\noindent\textbf{MSR3D}. We evaluated our model's performance against LLaVA-OV and LEO\footnote{The reported number reflects performance across all splits of MSR3D. However, since LEO was trained on ScanNet and 3RScan, its performance remains largely consistent across splits.}~\cite{huang2023embodied} on the Scannet split of MSR3D. Importantly, all comparisons are conducted in a zero-shot setting, as none of the models have been trained on MSR3D. This is an interleaved dataset where inputs consist of both text and images. However, since all models are tested in a zero-shot manner, they rely solely on text inputs, where the agent’s state descriptions and the question are combined into a single text prompt. On the visual processing side, LEO operates on 3D point cloud inputs, whereas both LLaVA-OV and our model rely only on RGB images. 

The results, presented in Table~\ref{tab:msqa}, demonstrate that our base model outperforms LLaVA-OV across all question types. Notably, aside from the "counting" category, our base model achieves approximately twice the performance of LLaVA-OV on other question types. Both models significantly outperform LEO, which struggles due to the absence of fine-tuning on this dataset.

Interestingly, we observe a slight performance drop when comparing SplatTalk with SplatTalk-ScanQA-FT. However, this drop becomes much more pronounced when compared with SplatTalk-3DVQA-FT. A plausible explanation is that fine-tuning on other QA datasets introduces overfitting, causing the model to adapt too specifically to those datasets. Furthermore, the observed performance drop suggests that MSR3D differs significantly from ScanQA and SQA3D, making direct knowledge transfer less effective. This distinction may also explain why LEO struggles to generalize in a zero-shot setting. Despite being trained on multiple QA datasets, none of them closely resemble MSR3D, limiting its ability to adapt without fine-tuning.


\subsection{Ablation Studies}
\textbf{Gaussian Sampling Strategies.}
We explore various Gaussian sampling strategies for selecting tokens as inputs to the LLM, including: (1) Random Sampling---Gaussians are selected randomly across the scene.
(2) Point Density Adaptive Sampling---Gaussians are sampled proportionally to the density of 3D points.
(3) Furthest Point Sampling (FPS)---Gaussians are iteratively selected at locations maximally distant from previously chosen points.
(4) Entropy Adaptive Sampling---Gaussians are prioritized based on language feature entropy, selecting points with the highest uncertainty. 

We conduct experiments on the ScanQA validation set. The results, presented in Table~\ref{tab:sampling_method}, demonstrate that entropy-adaptive sampling outperforms other sampling strategies, followed by FPS. This finding highlights entropy-adaptive sampling as the most effective approach for selecting informative tokens, making it the optimal strategy for downstream 3D VQA tasks.


\noindent\textbf{Length of Visual Tokens.} We evaluate the performance of SplatTalk on ScanQA, SQA3D, and MSR3D under two different settings: (1) using only $1$ image-worth of visual tokens ($729$ tokens) and (2) using the full context length corresponding to $44$ images-worth of visual tokens ($32,076$ tokens). The results, presented in Table~\ref{tab:visual_length}, evaluate model performance using EM@1 and EM@1-Refined, as these metrics offer a reliable measure of how precisely model predictions align with ground truth answers, both in literal accuracy and permissible variations in phrasing. We employ entropy-adaptive sampling (see Sec.~\ref{sec:entropy_sampling}), which prioritizes high-information tokens and remains deterministic given a fixed set of visual tokens.

A key observation is that performance improved across all datasets as the number of input tokens increased, with MSR3D showing the largest gain, nearly doubling in EM@1. In contrast, ScanQA and SQA3D saw only marginal improvements, even with significantly more tokens. This suggests that additional scene context may not substantially improve performance, possibly due to: (1) Dataset limitations where the tasks may not require rich spatial context; or (2) Model limitations where the model may be unable to effectively leverage extra visual tokens for spatial reasoning.


%% file: table/combine_ablation.tex
\begin{table*}[t]
\begin{minipage}{0.5\textwidth}

\input{table/sampling_ablation_rebuttal}

\end{minipage}
\hspace{5pt}
\begin{minipage}{0.5\textwidth}
\input{table/token_length_ablation}
\end{minipage}
\vspace{-10pt}
\end{table*}

%% file: table/sampling_ablation_rebuttal.tex
\begin{center}
\vspace{-10pt}

\begingroup
\setlength{\tabcolsep}{3pt} 
\renewcommand{\arraystretch}{1.2} 
\caption{\textbf{Ablation Study on Gaussian Sampling Strategies.} Our entropy-based sampling strategy consistently outperforms across multiple metrics.}
\vspace{-10pt}
\scalebox{0.6}{

 \begin{tabular}{l|c c c c c c}
        \toprule
        Sampling Method  & EM@1 & METEOR & ROUGE  & SPICE & Avg\\
        \midrule
        Random & 17.0  & 13.8  & \textbf{62.1} & 14.0 & 26.73\\
        Point Density & 16.2 &  \underline{14.0}  & 59.7 & \underline{14.1} & 26.00\\
        FPS & \textbf{17.7}  & 13.8  & \underline{62.0} & 13.6 & \underline{26.78}\\
        Entropy & \underline{17.1} &  \textbf{14.6}  & 61.7 & \textbf{14.2} & \textbf{26.90}\\
        \bottomrule

\end{tabular}
\label{tab:sampling_method}

}

\endgroup
\end{center}



%% file: table/token_length_ablation.tex
\begin{center}

\vspace{-10pt}
\begingroup
\setlength{\tabcolsep}{3pt} 
\renewcommand{\arraystretch}{1.2} 
\caption{\textbf{Ablation Study on Visual Input Length.} We compare model performance using $729$ tokens (equivalent to a single image) versus $32,076$ tokens (corresponding to $44$ images), analyzing the impact of visual context on reasoning.}
\vspace{-10pt}
\scalebox{0.7}{

 \begin{tabular}{l|c c |c c| c c}
        \toprule
        \multirow{2}{*}{Visual Length}& \multicolumn{2}{c|}{ScanQA (val)} & \multicolumn{2}{c|}{SQA3D (test)} & \multicolumn{2}{c}{MSR3D (test)}\\
          & EM@1 &EM@1-R &EM@1 &EM@1-R & EM@1 &EM@1-R\\
        \midrule
       $729$ tokens & 16.2 & 30.7 & 25.8 & 38.0 & 7.9 & 17.7 \\
        $32,076$  tokens & \textbf{17.1} & \textbf{32.2} & \textbf{26.1} & \textbf{41.3} & \textbf{14.1} & \textbf{23.3} \\
        \bottomrule

\end{tabular}

\label{tab:visual_length}
}
\endgroup

\end{center}



%% file: sec/5_conclusion.tex
\section{Conclusion}
In this work, we introduce the first method to integrate free-form language features into a 3D Gaussian Splatting framework, enabling zero-shot generalization for 3D Visual Question Answering (3D VQA) without requiring 3D point clouds or depth maps. Our approach surpasses methods that rely solely on 2D language feature maps and achieves performance on par with 3D LMMs that explicitly utilize 3D structural information across multiple 3D VQA benchmarks. We hope this work inspires further research into leveraging multi-view RGB inputs to encapsulate rich semantic representations of 3D scenes, ultimately enhancing 3D VQA capabilities without the reliance on explicit geometric priors.
\label{sec:conclusion}

\section*{Acknowledgment} We thank the creators of the datasets and codebases used in our work, ScanQA, SQA3D, MSR3D, ScanNet, LLaVA-OV, and FreeSplat, for making their resources publicly available. We are also grateful to the anonymous reviewers for their constructive feedback during the review process. Special thanks to Xiang Li and James M. Rehg for the insightful discussions.

%% file: sec/6_supp_mat.tex
\clearpage
\setcounter{page}{1}
\maketitlesupplementary


\section{Data Details}
We train SplatTalk on 500 ScanNet scenes from the ScanQA training set. This stage does not involve any language supervisory signal and is purely self-supervised. We evaluate our method and baselines on 3 datasets: ScanQA~\cite{azuma2022scanqa} validation set, SQA3D~\cite{ma2022sqa3d} test set, and MSR3D~\cite{linghu2025multi} test set. The ScanQA validation set comprises 71 ScanNet scenes, while the SQA3D and MSR3D test sets include 67 scenes, with significant overlap with ScanQA validation set.

\section{Training \& Model Details}

\noindent\textbf{SplatTalk Training.} The training of SplatTalk consists of the following main steps: 1) Extract high-dimensional language features for each RGB image using LLaVA-OV~\cite{li2024llava}, 2) Train an autoencoder to compress the high-dimensional features obtained from step 1 into a more compact space, and 3) Train the 3D Gaussian Splatting (3DGS) model.

First, to extract language-informed visual features, we leverage LLaVA-OV~\cite{li2024llava}, which processes RGB images through its SigLIP~\cite{Zhai2023SigLIP}-based vision encoder, followed by a multimodal projector. This transformation ensures that the extracted visual tokens are natively structured for the LLM’s input space, allowing seamless processing and comprehension by the underlying Qwen2~\cite{yang2024qwen2} model. These features serve as pseudo 2D ground truth, guiding the training of the 3D-language Gaussian Splatting model.

After obtaining these high-dimensional semantic features, we train an autoencoder to compress them while preserving their most informative components. Each feature map of size $27\times27\times3584$, where $3584$ represents the feature depth, is then flattened, resulting in $729$ feature vectors, each of size $1\times3584$.

The autoencoder is composed of multiple linear layers, each interleaved with a batch normalization (BN) layer and a GeLU activation function. Specifically,
\begin{align*}
\text{Encoder}: (3584) \rightarrow (2048) \rightarrow \text{BN} 
\rightarrow \text{GeLU} \rightarrow (1024) \\
 \rightarrow \text{BN} \rightarrow \text{GeLU} \rightarrow (512) \rightarrow  \text{BN} \rightarrow \text{GeLU} \rightarrow (256)
\end{align*}
\begin{align*}
\text{Decoder}: (256) \rightarrow (512) \rightarrow \text{GeLU} \rightarrow (1024) \rightarrow \text{GeLU}\\ \rightarrow (2048)
\rightarrow \text{GeLU} \rightarrow (2048) \rightarrow \text{GeLU} \rightarrow (3584)
\end{align*}
where $(3584)$ indicates the size of the linear layer, BN stands for Batch Norm and GeLU stands for the GeLU activation function. The compressed features are normalized to reside within a unit hypersphere, which we find to be important for stabilizing the 3DGS training process.

We train the autoencoder with batch size $256$, using AdamW~\cite{loshchilov2017decoupled} optimizer with learning rate $0.0001$ for $100$ epochs on a single NVIDIA H100 (80GB) GPU.

SplatTalk builds upon the FreeSplat~\cite{wang2025freesplat} architecture, extending it with a language feature head attached to the Gaussian latent decoder. This additional head predicts a $256$-dimensional feature vector for each Gaussian, alongside the standard Gaussian parameters used for RGB splatting.

For training the 3DGS component, we uniformly sample 100 views per scene. Each input image is resized to $32 \times 32$ resolution to balance efficiency and feature retention. The language features are trained using compressed pseudo 2D ground truth as supervisory signal. Inspired by~\cite{zhou2024feature}, we implement a parallel CUDA rasterizer pipeline, allowing RGB and language features to be rendered in parallel using the shared Gaussian parameters. This approach is memory efficient and allows for rendering higher-dimensional features than the traditional rendering pipeline. 

SplatTalk is trained using the Adam optimizer with an initial learning rate of $1e-4$, which decays smoothly following a cosine decay schedule. We employ a combination of photometric and semantic losses to effectively train SplatTalk. The photometric loss is computed using a combination of MSE and LPIPs, and semantic loss using MSE and cosine distance. The final loss is
\[
\mathcal{L}=\Vert I-\hat{I}\Vert^2 + 0.05\cdot LPIPS + \Vert F-\hat{F}\Vert^2 + 1-\cos(F,\hat{F})
\]
where $I$ and $F$ are the ground-truth RGB image and the pseudo ground-truth language feature map respectively. $\hat{I}$ and $\hat{F}$ correspond to the predicted RGB image and language feature map respectively. We train SplatTalk on one NVIDIA H100 (80GB) GPU.

\noindent\textbf{SplatTalk Inference.} We begin by extracting the reconstructed 3D Gaussians that represent the entire scene. Next, we retrieve the language features corresponding to each 3D Gaussian, capturing the semantic information embedded within the scene representation.

\noindent\textbf{3D VQA Inference.} We sample $32,076$ tokens from the set of Gaussian tokens obtained from the SplatTalk inference step. This is equivalent to $44$ image tokens, which is the context window size of the LLM. The optimal sampling method we use is entropy-based sampling. We first compute the entropy for each 3D Gaussian. Then we rank them in descending order from highest to lowest entropy. Finally, we sample the top $32,076$ tokens from this list as the input visual tokens to the LLM, along with the language tokens.

\noindent\textbf{3D VQA Fine-tuning.} During fine-tuning we employ LoRA~\cite{hu2021lora} with $r=16$ and $\alpha=64$. The model is fine-tuned for $1$ epoch with learning rate $1e-5$ on a single NVIDIA H100 (80GB) GPU.

\section{Evaluation Metrics}
We evaluate model performance on ScanQA using n-gram-based metrics, including CIDEr~\cite{vedantam2015cider}, METEOR~\cite{banerjee2005meteor}, ROUGE~\cite{lin2004rouge}, EM@1~\cite{goyal2017making}, and EM@1-Refined. For SQA3D, we report EM@1 and EM@1-Refined, while for MSR3D, we follow the authors' recommendation and use the LLM-Match metric~\cite{Liu2024MMBench} to assess answer quality.

Specifically, CIDEr evaluates the generated response by comparing n-gram similarity against multiple reference responses, assigning higher weight to words that frequently appear across multiple references. In contrast, METEOR aligns words between the generated response and reference answers using exact matches, stemming, synonyms, and paraphrasing, providing a more flexible and semantically aware evaluation. It computes both precision and recall, with a higher emphasis on recall, as human evaluation favors covering all key reference words. METEOR captures both semantic and syntactic similarity, making it more robust than purely n-gram-based metrics. ROUGE-L, on the other hand, measures similarity using the Longest Common Subsequence (LCS) between the generated response and the reference answers.

On the other hand, EM@1 or Exact Match at top-1, is a strict evaluation metric that only assigns a score if the generated response perfectly matches the reference answer, without any deviations. Even if the response is semantically correct but differs in phrasing, formatting, or minor details, the model is penalized. In contrast, EM@1-Refined offers a more flexible evaluation, allowing for slight variations in wording.

The LLM-Match metric, introduced in~\cite{Liu2024MMBench}, uses GPT-4 to assign a score from $1$ to $5$ to the generated responses based on their similarity to the reference answers. A score of $1$ indicates a completely incorrect response, while a score of $5$ corresponds to a fully correct answer. Intermediate scores reflect varying degrees of alignment.

\section{More Visualization on ScanQA}
\begin{figure*}[t!]
\centering
\includegraphics[width=\linewidth]{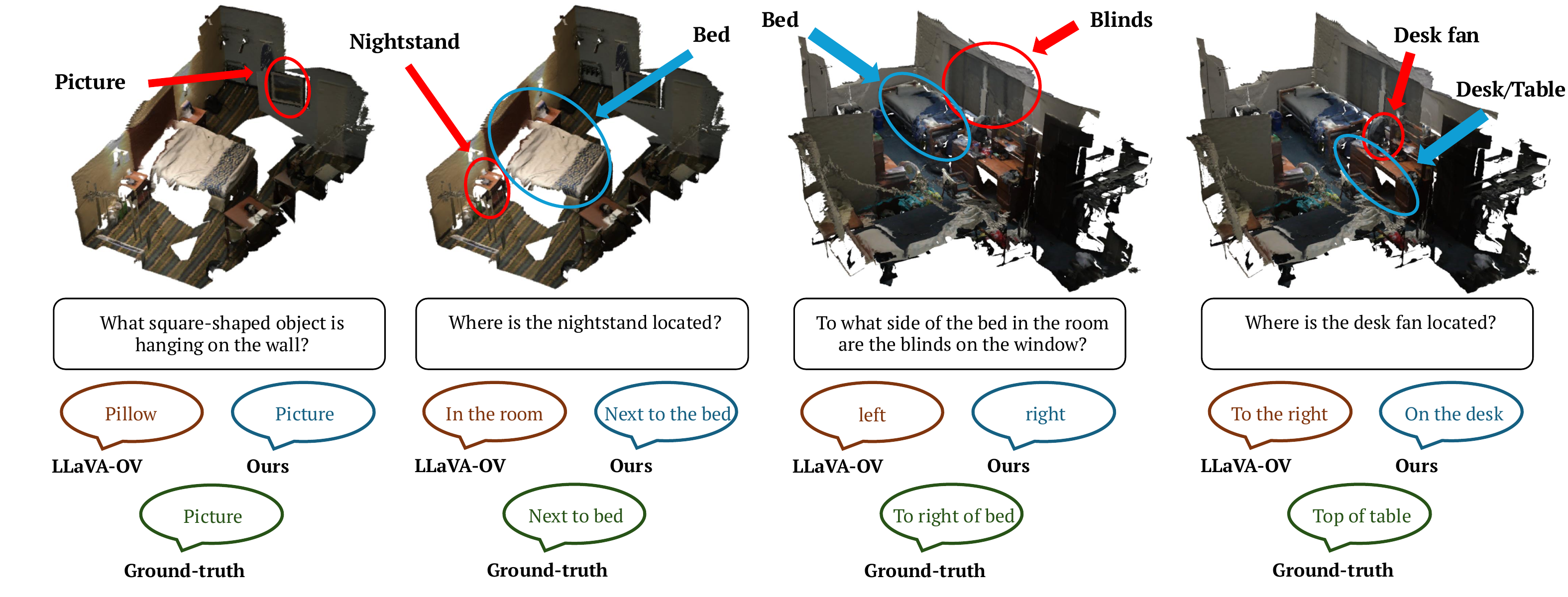}
\caption{\textbf{Qualitative results on ScanQA}, scene0389\_00 and scene0222\_00. We compare responses from LLaVA-OV, our model (Ours), and the ground truth (GT) for spatial reasoning questions in 3D VQA. Each scene highlights the referenced objects with red circles, and key relative objects are marked in blue. The answers from each model are displayed in color-coded speech bubbles: LLaVA-OV (brown), Ours (blue), and GT (green). With its 3D-aware representation, our model exhibits improved spatial reasoning capabilities.}
\label{fig:example_viz_sup}
\end{figure*}

\begin{figure*}[t!]
\centering
\includegraphics[width=\linewidth]{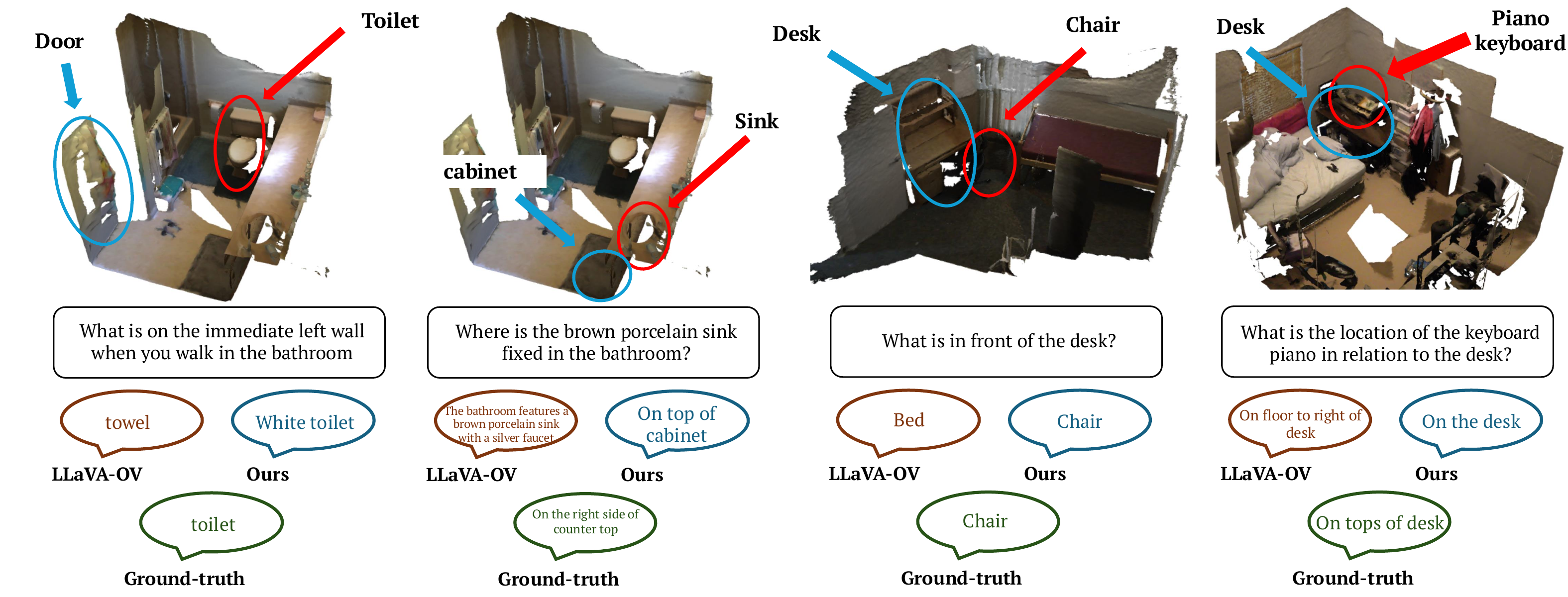}
\caption{\textbf{Qualitative results on ScanQA}, scene0100\_00, scene0193\_00, and scene0426\_00. We compare responses from LLaVA-OV, our model (Ours), and the ground truth (GT) for spatial reasoning questions in 3D VQA. Each scene highlights the referenced objects with red circles, and key relative objects are marked in blue. The answers from each model are displayed in color-coded speech bubbles: LLaVA-OV (brown), Ours (blue), and GT (green). With its 3D-aware representation, our model exhibits improved spatial reasoning capabilities.}
\label{fig:example_viz_sup3}

\end{figure*}

In Fig.~\ref{fig:example_viz_sup} and Fig.~\ref{fig:example_viz_sup3}, we show more qualitative examples on ScanQA. Our SplatTalk can reason about the spatial relationships between objects in the scene significantly better than LLaVA-OV in many cases.

%% file: main.bbl
\begin{thebibliography}{53}
\providecommand{\natexlab}[1]{#1}
\providecommand{\url}[1]{\texttt{#1}}
\expandafter\ifx\csname urlstyle\endcsname\relax
  \providecommand{\doi}[1]{doi: #1}\else
  \providecommand{\doi}{doi: \begingroup \urlstyle{rm}\Url}\fi

\bibitem[Anthropic(2024)]{anthropic2024claude3}
Anthropic.
\newblock Claude 3.
\newblock \url{https://www.anthropic.com/index/claude}, 2024.
\newblock Accessed: 2024-02-28.

\bibitem[Azuma et~al.(2022)Azuma, Miyanishi, Kurita, and Kawanabe]{azuma2022scanqa}
Daichi Azuma, Taiki Miyanishi, Shuhei Kurita, and Motoaki Kawanabe.
\newblock Scanqa: 3d question answering for spatial scene understanding.
\newblock In \emph{proceedings of the IEEE/CVF conference on computer vision and pattern recognition}, pages 19129--19139, 2022.

\bibitem[Banerjee and Lavie(2005)]{banerjee2005meteor}
Satanjeev Banerjee and Alon Lavie.
\newblock {METEOR: An Automatic Metric for MT Evaluation with Improved Correlation with Human Judgments}.
\newblock In \emph{Proceedings of the ACL Workshop on Intrinsic and Extrinsic Evaluation Measures for Machine Translation and/or Summarization}, pages 65--72, 2005.

\bibitem[Bhalgat et~al.(2024)Bhalgat, Laina, Henriques, Zisserman, and Vedaldi]{bhalgat2024n2f2}
Yash Bhalgat, Iro Laina, Jo{\~a}o~F Henriques, Andrew Zisserman, and Andrea Vedaldi.
\newblock N2f2: Hierarchical scene understanding with nested neural feature fields.
\newblock In \emph{European Conference on Computer Vision}, pages 197--214. Springer, 2024.

\bibitem[Charatan et~al.(2024)Charatan, Li, Tagliasacchi, and Sitzmann]{charatan2024pixelsplat}
David Charatan, Sizhe~Lester Li, Andrea Tagliasacchi, and Vincent Sitzmann.
\newblock pixelsplat: 3d gaussian splats from image pairs for scalable generalizable 3d reconstruction.
\newblock In \emph{Proceedings of the IEEE/CVF conference on computer vision and pattern recognition}, pages 19457--19467, 2024.

\bibitem[Chen et~al.(2024{\natexlab{a}})Chen, Wei, and Lee]{chen2024chatsplat}
Hanlin Chen, Fangyin Wei, and Gim~Hee Lee.
\newblock Chatsplat: 3d conversational gaussian splatting.
\newblock \emph{arXiv preprint arXiv:2412.00734}, 2024{\natexlab{a}}.

\bibitem[Chen et~al.(2024{\natexlab{b}})Chen, Chen, Zhang, Li, Yu, Fei, Zhu, Fan, and Chen]{chen2024ll3da}
Sijin Chen, Xin Chen, Chi Zhang, Mingsheng Li, Gang Yu, Hao Fei, Hongyuan Zhu, Jiayuan Fan, and Tao Chen.
\newblock Ll3da: Visual interactive instruction tuning for omni-3d understanding reasoning and planning.
\newblock In \emph{Proceedings of the IEEE/CVF Conference on Computer Vision and Pattern Recognition}, pages 26428--26438, 2024{\natexlab{b}}.

\bibitem[Chen et~al.(2024{\natexlab{c}})Chen, Xu, Zheng, Zhuang, Pollefeys, Geiger, Cham, and Cai]{chen2024mvsplat}
Yuedong Chen, Haofei Xu, Chuanxia Zheng, Bohan Zhuang, Marc Pollefeys, Andreas Geiger, Tat-Jen Cham, and Jianfei Cai.
\newblock Mvsplat: Efficient 3d gaussian splatting from sparse multi-view images.
\newblock In \emph{European Conference on Computer Vision}, pages 370--386. Springer, 2024{\natexlab{c}}.

\bibitem[Chen et~al.(2024{\natexlab{d}})Chen, Yang, Huang, Wang, Xu, Lyu, Lin, and Pang]{chen2024grounded}
Yilun Chen, Shuai Yang, Haifeng Huang, Tai Wang, Runsen Xu, Ruiyuan Lyu, Dahua Lin, and Jiangmiao Pang.
\newblock Grounded 3d-llm with referent tokens.
\newblock \emph{arXiv preprint arXiv:2405.10370}, 2024{\natexlab{d}}.

\bibitem[Engelmann et~al.(2024)Engelmann, Manhardt, Niemeyer, Tateno, Pollefeys, and Tombari]{engelmann2024opennerf}
Francis Engelmann, Fabian Manhardt, Michael Niemeyer, Keisuke Tateno, Marc Pollefeys, and Federico Tombari.
\newblock Opennerf: open set 3d neural scene segmentation with pixel-wise features and rendered novel views.
\newblock \emph{arXiv preprint arXiv:2404.03650}, 2024.

\bibitem[et~al.(2024{\natexlab{a}})]{yang2024qwen2}
An~Yang et al.
\newblock {Qwen2} technical report.
\newblock \emph{arXiv preprint arXiv:2407.10671}, 2024{\natexlab{a}}.

\bibitem[et~al.(2024{\natexlab{b}})]{geminiteam2024geminifamilyhighlycapable}
Gemini~Team et al.
\newblock Gemini: A family of highly capable multimodal models, 2024{\natexlab{b}}.

\bibitem[Fan et~al.(2025)Fan, Zhang, Cong, Wang, Li, Wen, Zhou, Kadambi, Wang, Xu, et~al.]{fan2025large}
Zhiwen Fan, Jian Zhang, Wenyan Cong, Peihao Wang, Renjie Li, Kairun Wen, Shijie Zhou, Achuta Kadambi, Zhangyang Wang, Danfei Xu, et~al.
\newblock Large spatial model: End-to-end unposed images to semantic 3d.
\newblock \emph{Advances in Neural Information Processing Systems}, 37:\penalty0 40212--40229, 2025.

\bibitem[Fu et~al.(2024)Fu, Liu, Chen, Nie, and Xiong]{fu2024scene}
Rao Fu, Jingyu Liu, Xilun Chen, Yixin Nie, and Wenhan Xiong.
\newblock Scene-llm: Extending language model for 3d visual understanding and reasoning.
\newblock \emph{arXiv preprint arXiv:2403.11401}, 2024.

\bibitem[Goyal et~al.(2017)Goyal, Khot, Summers-Stay, Batra, and Parikh]{goyal2017making}
Yash Goyal, Tejas Khot, Douglas Summers-Stay, Dhruv Batra, and Devi Parikh.
\newblock {Making the V in VQA Matter: Elevating the Role of Image Understanding in Visual Question Answering}.
\newblock In \emph{Proceedings of the IEEE Conference on Computer Vision and Pattern Recognition (CVPR)}, pages 6904--6913, 2017.

\bibitem[Hong et~al.(2023)Hong, Zhen, Chen, Zheng, Du, Chen, and Gan]{hong20233d}
Yining Hong, Haoyu Zhen, Peihao Chen, Shuhong Zheng, Yilun Du, Zhenfang Chen, and Chuang Gan.
\newblock 3d-llm: Injecting the 3d world into large language models.
\newblock \emph{Advances in Neural Information Processing Systems}, 36:\penalty0 20482--20494, 2023.

\bibitem[Hu et~al.(2022)Hu, Shen, Wallis, Allen-Zhu, Li, Wang, Wang, and Chen]{hu2021lora}
Edward~J. Hu, Yelong Shen, Phillip Wallis, Zeyuan Allen-Zhu, Yuanzhi Li, Shean Wang, Lu Wang, and Weizhu Chen.
\newblock {LoRA}: Low-rank adaptation of large language models.
\newblock In \emph{International Conference on Learning Representations (ICLR)}, 2022.

\bibitem[Huang et~al.(2023{\natexlab{a}})Huang, Chen, Wang, Huang, Xu, Wang, Liu, Cheng, Zhao, Pang, et~al.]{huang2023chat}
Haifeng Huang, Yilun Chen, Zehan Wang, Rongjie Huang, Runsen Xu, Tai Wang, Luping Liu, Xize Cheng, Yang Zhao, Jiangmiao Pang, et~al.
\newblock Chat-scene: Bridging 3d scene and large language models with object identifiers.
\newblock \emph{arXiv preprint arXiv:2312.08168}, 2023{\natexlab{a}}.

\bibitem[Huang et~al.(2023{\natexlab{b}})Huang, Yong, Ma, Linghu, Li, Wang, Li, Zhu, Jia, and Huang]{huang2023embodied}
Jiangyong Huang, Silong Yong, Xiaojian Ma, Xiongkun Linghu, Puhao Li, Yan Wang, Qing Li, Song-Chun Zhu, Baoxiong Jia, and Siyuan Huang.
\newblock An embodied generalist agent in 3d world.
\newblock \emph{arXiv preprint arXiv:2311.12871}, 2023{\natexlab{b}}.

\bibitem[Kang et~al.(2025)Kang, Huang, Shang, Shah, and Yan]{kang2025robin3dimproving3dlarge}
Weitai Kang, Haifeng Huang, Yuzhang Shang, Mubarak Shah, and Yan Yan.
\newblock Robin3d: Improving 3d large language model via robust instruction tuning, 2025.

\bibitem[Kerr et~al.(2023)Kerr, Kim, Goldberg, Kanazawa, and Tancik]{kerr2023lerf}
Justin Kerr, Chung~Min Kim, Ken Goldberg, Angjoo Kanazawa, and Matthew Tancik.
\newblock Lerf: Language embedded radiance fields.
\newblock In \emph{Proceedings of the IEEE/CVF International Conference on Computer Vision}, pages 19729--19739, 2023.

\bibitem[Kirillov et~al.(2023)Kirillov, Mintun, Ravi, Mao, Rolland, Gustafson, Xiao, Whitehead, Berg, Lo, et~al.]{kirillov2023segment}
Alexander Kirillov, Eric Mintun, Nikhila Ravi, Hanzi Mao, Chloe Rolland, Laura Gustafson, Tete Xiao, Spencer Whitehead, Alexander~C Berg, Wan-Yen Lo, et~al.
\newblock Segment anything.
\newblock In \emph{Proceedings of the IEEE/CVF international conference on computer vision}, pages 4015--4026, 2023.

\bibitem[Li et~al.(2024{\natexlab{a}})Li, Zhang, Guo, Zhang, Li, Zhang, Zhang, Zhang, Li, Liu, et~al.]{li2024llava}
Bo Li, Yuanhan Zhang, Dong Guo, Renrui Zhang, Feng Li, Hao Zhang, Kaichen Zhang, Peiyuan Zhang, Yanwei Li, Ziwei Liu, et~al.
\newblock Llava-onevision: Easy visual task transfer.
\newblock \emph{arXiv preprint arXiv:2408.03326}, 2024{\natexlab{a}}.

\bibitem[Li et~al.(2024{\natexlab{b}})Li, Zhang, Zhang, Zhang, Li, Li, Ma, and Li]{li2024llavanextinterleavetacklingmultiimagevideo}
Feng Li, Renrui Zhang, Hao Zhang, Yuanhan Zhang, Bo Li, Wei Li, Zejun Ma, and Chunyuan Li.
\newblock Llava-next-interleave: Tackling multi-image, video, and 3d in large multimodal models, 2024{\natexlab{b}}.

\bibitem[Li et~al.(2024{\natexlab{c}})Li, Wu, Meng, Gao, Zhang, Wang, and Zhang]{li2024instancegaussian}
Haijie Li, Yanmin Wu, Jiarui Meng, Qiankun Gao, Zhiyao Zhang, Ronggang Wang, and Jian Zhang.
\newblock Instancegaussian: Appearance-semantic joint gaussian representation for 3d instance-level perception.
\newblock \emph{arXiv preprint arXiv:2411.19235}, 2024{\natexlab{c}}.

\bibitem[Li et~al.(2023)Li, Li, Savarese, and Hoi]{li2023blip}
Junnan Li, Dongxu Li, Silvio Savarese, and Steven Hoi.
\newblock Blip-2: Bootstrapping language-image pre-training with frozen image encoders and large language models.
\newblock In \emph{International conference on machine learning}, pages 19730--19742. PMLR, 2023.

\bibitem[Li et~al.(2024{\natexlab{d}})Li, Wang, He, Li, Wang, Liu, Wang, Xu, Chen, Luo, et~al.]{li2024mvbench}
Kunchang Li, Yali Wang, Yinan He, Yizhuo Li, Yi Wang, Yi Liu, Zun Wang, Jilan Xu, Guo Chen, Ping Luo, et~al.
\newblock Mvbench: A comprehensive multi-modal video understanding benchmark.
\newblock In \emph{Proceedings of the IEEE/CVF Conference on Computer Vision and Pattern Recognition}, pages 22195--22206, 2024{\natexlab{d}}.

\bibitem[Lin(2004)]{lin2004rouge}
Chin-Yew Lin.
\newblock {ROUGE: A Package for Automatic Evaluation of Summaries}.
\newblock In \emph{Proceedings of the Workshop on Text Summarization Branches Out}, pages 74--81, 2004.

\bibitem[Linghu et~al.(2025)Linghu, Huang, Niu, Ma, Jia, and Huang]{linghu2025multi}
Xiongkun Linghu, Jiangyong Huang, Xuesong Niu, Xiaojian~Shawn Ma, Baoxiong Jia, and Siyuan Huang.
\newblock Multi-modal situated reasoning in 3d scenes.
\newblock \emph{Advances in Neural Information Processing Systems}, 37:\penalty0 140903--140936, 2025.

\bibitem[Liu et~al.(2023)Liu, Li, Wu, and Lee]{liu2023visual}
Haotian Liu, Chunyuan Li, Qingyang Wu, and Yong~Jae Lee.
\newblock Visual instruction tuning.
\newblock \emph{Advances in neural information processing systems}, 36:\penalty0 34892--34916, 2023.

\bibitem[Liu et~al.(2024)Liu, Duan, Zhang, Li, Zhang, Zhao, Yuan, Wang, He, Liu, Chen, and Lin]{Liu2024MMBench}
Yuan Liu, Haodong Duan, Yuanhan Zhang, Bo Li, Songyang Zhang, Wangbo Zhao, Yike Yuan, Jiaqi Wang, Conghui He, Ziwei Liu, Kai Chen, and Dahua Lin.
\newblock {MMBench}: Is your multi-modal model an all-around player?
\newblock In \emph{Proceedings of the European Conference on Computer Vision (ECCV)}, 2024.
\newblock Accepted as Oral Presentation.

\bibitem[Loshchilov and Hutter(2017)]{loshchilov2017decoupled}
Ilya Loshchilov and Frank Hutter.
\newblock Decoupled weight decay regularization.
\newblock \emph{arXiv preprint arXiv:1711.05101}, 2017.

\bibitem[Lu et~al.(2024)Lu, Liu, Zhang, Wang, Dong, Liu, Sun, Ren, Li, Yang, Sun, Deng, Xu, Xie, and Ruan]{lu2024deepseekvl}
Haoyu Lu, Wen Liu, Bo Zhang, Bingxuan Wang, Kai Dong, Bo Liu, Jingxiang Sun, Tongzheng Ren, Zhuoshu Li, Hao Yang, Yaofeng Sun, Chengqi Deng, Hanwei Xu, Zhenda Xie, and Chong Ruan.
\newblock Deepseek-vl: Towards real-world vision-language understanding, 2024.

\bibitem[Ma et~al.(2022)Ma, Yong, Zheng, Li, Liang, Zhu, and Huang]{ma2022sqa3d}
Xiaojian Ma, Silong Yong, Zilong Zheng, Qing Li, Yitao Liang, Song-Chun Zhu, and Siyuan Huang.
\newblock Sqa3d: Situated question answering in 3d scenes.
\newblock \emph{arXiv preprint arXiv:2210.07474}, 2022.

\bibitem[OpenAI(2023)]{openai2023gpt4v}
OpenAI.
\newblock Gpt-4 with vision (gpt-4v).
\newblock \url{https://openai.com/research/gpt-4}, 2023.
\newblock Accessed: 2024-02-28.

\bibitem[Peng et~al.(2024)Peng, Planche, Gao, Zheng, Choudhuri, Chen, Chen, and Wu]{peng20243d}
Qucheng Peng, Benjamin Planche, Zhongpai Gao, Meng Zheng, Anwesa Choudhuri, Terrence Chen, Chen Chen, and Ziyan Wu.
\newblock 3d vision-language gaussian splatting.
\newblock \emph{arXiv preprint arXiv:2410.07577}, 2024.

\bibitem[Peng et~al.(2023)Peng, Genova, Jiang, Tagliasacchi, Pollefeys, Funkhouser, et~al.]{peng2023openscene}
Songyou Peng, Kyle Genova, Chiyu Jiang, Andrea Tagliasacchi, Marc Pollefeys, Thomas Funkhouser, et~al.
\newblock Openscene: 3d scene understanding with open vocabularies.
\newblock In \emph{Proceedings of the IEEE/CVF conference on computer vision and pattern recognition}, pages 815--824, 2023.

\bibitem[Qin et~al.(2024)Qin, Li, Zhou, Wang, and Pfister]{qin2024langsplat}
Minghan Qin, Wanhua Li, Jiawei Zhou, Haoqian Wang, and Hanspeter Pfister.
\newblock Langsplat: 3d language gaussian splatting.
\newblock In \emph{Proceedings of the IEEE/CVF Conference on Computer Vision and Pattern Recognition}, pages 20051--20060, 2024.

\bibitem[Radford et~al.(2021)Radford, Kim, Hallacy, Ramesh, Goh, Agarwal, Sastry, Askell, Mishkin, Clark, et~al.]{radford2021learning}
Alec Radford, Jong~Wook Kim, Chris Hallacy, Aditya Ramesh, Gabriel Goh, Sandhini Agarwal, Girish Sastry, Amanda Askell, Pamela Mishkin, Jack Clark, et~al.
\newblock Learning transferable visual models from natural language supervision.
\newblock In \emph{International conference on machine learning}, pages 8748--8763. PmLR, 2021.

\bibitem[Rasheed et~al.(2024)Rasheed, Maaz, Shaji, Shaker, Khan, Cholakkal, Anwer, Xing, Yang, and Khan]{rasheed2024glamm}
Hanoona Rasheed, Muhammad Maaz, Sahal Shaji, Abdelrahman Shaker, Salman Khan, Hisham Cholakkal, Rao~M Anwer, Eric Xing, Ming-Hsuan Yang, and Fahad~S Khan.
\newblock Glamm: Pixel grounding large multimodal model.
\newblock In \emph{Proceedings of the IEEE/CVF Conference on Computer Vision and Pattern Recognition}, pages 13009--13018, 2024.

\bibitem[Vedantam et~al.(2015)Vedantam, Zitnick, and Parikh]{vedantam2015cider}
Ramakrishna Vedantam, C.~Lawrence Zitnick, and Devi Parikh.
\newblock {CIDEr: Consensus-based Image Description Evaluation}.
\newblock In \emph{Proceedings of the IEEE Conference on Computer Vision and Pattern Recognition (CVPR)}, pages 4566--4575, 2015.

\bibitem[Wang et~al.(2024{\natexlab{a}})Wang, Fan, Wang, Su, Ramamoorthi, et~al.]{wang2024lift3d}
Peihao Wang, Zhiwen Fan, Zhangyang Wang, Hao Su, Ravi Ramamoorthi, et~al.
\newblock Lift3d: Zero-shot lifting of any 2d vision model to 3d.
\newblock In \emph{Proceedings of the IEEE/CVF Conference on Computer Vision and Pattern Recognition}, pages 21367--21377, 2024{\natexlab{a}}.

\bibitem[Wang et~al.(2025{\natexlab{a}})Wang, Lv, Yu, Hong, Qi, Wang, Ji, Yang, Zhao, XiXuan, et~al.]{wang2025cogvlm}
Weihan Wang, Qingsong Lv, Wenmeng Yu, Wenyi Hong, Ji Qi, Yan Wang, Junhui Ji, Zhuoyi Yang, Lei Zhao, Song XiXuan, et~al.
\newblock Cogvlm: Visual expert for pretrained language models.
\newblock \emph{Advances in Neural Information Processing Systems}, 37:\penalty0 121475--121499, 2025{\natexlab{a}}.

\bibitem[Wang et~al.(2023{\natexlab{a}})Wang, Ma, Li, Kortylewski, and Yuille]{wang20233d}
Xingrui Wang, Wufei Ma, Zhuowan Li, Adam Kortylewski, and Alan~L Yuille.
\newblock 3d-aware visual question answering about parts, poses and occlusions.
\newblock \emph{Advances in Neural Information Processing Systems}, 36:\penalty0 58717--58735, 2023{\natexlab{a}}.

\bibitem[Wang et~al.(2024{\natexlab{b}})Wang, Lan, Zhu, Chen, and Lu]{wang2024gsemsplatgeneralizablesemantic3d}
Xingrui Wang, Cuiling Lan, Hanxin Zhu, Zhibo Chen, and Yan Lu.
\newblock Gsemsplat: Generalizable semantic 3d gaussian splatting from uncalibrated image pairs, 2024{\natexlab{b}}.

\bibitem[Wang et~al.(2025{\natexlab{b}})Wang, Huang, Chen, and Lee]{wang2025freesplat}
Yunsong Wang, Tianxin Huang, Hanlin Chen, and Gim~Hee Lee.
\newblock Freesplat: Generalizable 3d gaussian splatting towards free view synthesis of indoor scenes.
\newblock \emph{Advances in Neural Information Processing Systems}, 37:\penalty0 107326--107349, 2025{\natexlab{b}}.

\bibitem[Wang et~al.(2023{\natexlab{b}})Wang, Huang, Zhao, Zhang, and Zhao]{wang2023chat}
Zehan Wang, Haifeng Huang, Yang Zhao, Ziang Zhang, and Zhou Zhao.
\newblock Chat-3d: Data-efficiently tuning large language model for universal dialogue of 3d scenes.
\newblock \emph{arXiv preprint arXiv:2308.08769}, 2023{\natexlab{b}}.

\bibitem[Wu et~al.(2025)Wu, Meng, Li, Wu, Shi, Cheng, Zhao, Feng, Ding, Wang, et~al.]{wu2025opengaussian}
Yanmin Wu, Jiarui Meng, Haijie Li, Chenming Wu, Yahao Shi, Xinhua Cheng, Chen Zhao, Haocheng Feng, Errui Ding, Jingdong Wang, et~al.
\newblock Opengaussian: Towards point-level 3d gaussian-based open vocabulary understanding.
\newblock \emph{Advances in Neural Information Processing Systems}, 37:\penalty0 19114--19138, 2025.

\bibitem[Yu et~al.(2019)Yu, Yu, Cui, Tao, and Tian]{yu2019deep}
Zhou Yu, Jun Yu, Yuhao Cui, Dacheng Tao, and Qi Tian.
\newblock Deep modular co-attention networks for visual question answering.
\newblock In \emph{Proceedings of the IEEE/CVF conference on computer vision and pattern recognition}, pages 6281--6290, 2019.

\bibitem[Zhai et~al.(2023)Zhai, Mustafa, Kolesnikov, and Beyer]{Zhai2023SigLIP}
Xiaohua Zhai, Basil Mustafa, Alexander Kolesnikov, and Lucas Beyer.
\newblock {Sigmoid Loss for Language Image Pre-Training}.
\newblock In \emph{Proceedings of the IEEE/CVF International Conference on Computer Vision (ICCV)}, 2023.

\bibitem[Zhou et~al.(2024)Zhou, Chang, Jiang, Fan, Zhu, Xu, Chari, You, Wang, and Kadambi]{zhou2024feature}
Shijie Zhou, Haoran Chang, Sicheng Jiang, Zhiwen Fan, Zehao Zhu, Dejia Xu, Pradyumna Chari, Suya You, Zhangyang Wang, and Achuta Kadambi.
\newblock Feature 3dgs: Supercharging 3d gaussian splatting to enable distilled feature fields.
\newblock In \emph{Proceedings of the IEEE/CVF Conference on Computer Vision and Pattern Recognition}, pages 21676--21685, 2024.

\bibitem[Zhu et~al.(2024)Zhu, Wang, Zhang, Pang, and Liu]{zhu2024llava}
Chenming Zhu, Tai Wang, Wenwei Zhang, Jiangmiao Pang, and Xihui Liu.
\newblock Llava-3d: A simple yet effective pathway to empowering lmms with 3d-awareness.
\newblock \emph{arXiv preprint arXiv:2409.18125}, 2024.

\bibitem[Zhu et~al.(2023)Zhu, Ma, Chen, Deng, Huang, and Li]{zhu20233d}
Ziyu Zhu, Xiaojian Ma, Yixin Chen, Zhidong Deng, Siyuan Huang, and Qing Li.
\newblock 3d-vista: Pre-trained transformer for 3d vision and text alignment.
\newblock In \emph{Proceedings of the IEEE/CVF International Conference on Computer Vision}, pages 2911--2921, 2023.

\end{thebibliography}
